\documentclass{article}

\usepackage{microtype}
\usepackage{graphicx}
\usepackage{subfigure}
\usepackage{booktabs} 

\usepackage{hyperref}

\usepackage[accepted]{icml2025}

\usepackage{amsmath}
\usepackage{amssymb}
\usepackage{mathtools}
\usepackage{amsthm}

\usepackage[capitalize,noabbrev]{cleveref}

\theoremstyle{plain}

\theoremstyle{definition}

\theoremstyle{remark}

\usepackage[textsize=tiny]{todonotes}

\icmltitlerunning{Deep Unfolding with Kernel-based Quantization in MIMO Detection}

\begin{document}

\twocolumn[
\icmltitle{Deep Unfolding with Kernel-based Quantization in MIMO Detection}



\icmlsetsymbol{equal}{*}

\begin{icmlauthorlist}
\icmlauthor{Zeyi Ren}{hku}
\icmlauthor{Jingreng Lei}{hku}
\icmlauthor{Yichen Jin}{hku}
\icmlauthor{Ermo Hua}{thu}
\icmlauthor{Qingfeng Lin}{hku}
\icmlauthor{Chen Zhang}{hku}
\icmlauthor{Bowen Zhou}{thu,ailab}
\icmlauthor{Yik-Chung Wu}{hku}

\end{icmlauthorlist}

\icmlaffiliation{hku}{Department of Electrical and Electronic Engineering, The University of Hong Kong, HKSAR, China}
\icmlaffiliation{thu}{Tsinghua University}
\icmlaffiliation{ailab}{Shanghai AI Laboratory}

\icmlcorrespondingauthor{Yik-Chung Wu}{ycwu@eee.hku.hk}

\icmlkeywords{Nonparametric quantization,
deep unfolding,
kernel methods,
edge intelligence.}

\vskip 0.3in
]



\printAffiliationsAndNotice{} 

\begin{abstract}
The development of edge computing places critical demands on energy-efficient model deployment for multiple-input multiple-output (MIMO) detection tasks. Deploying deep unfolding models such as PGD-Nets and ADMM-Nets into resource-constrained edge devices using quantization methods is challenging. Existing quantization methods based on quantization aware training (QAT) suffer from performance degradation due to their reliance on parametric distribution assumption of activations and static quantization step sizes. To address these challenges, this paper proposes a novel kernel-based adaptive quantization (KAQ) framework for deep unfolding networks. By utilizing a joint kernel density estimation (KDE) and maximum mean discrepancy (MMD) approach to align activation distributions between full-precision and quantized models, the need for prior distribution assumptions is eliminated. Additionally, a dynamic step size updating method is introduced to adjust the quantization step size based on the channel conditions of wireless networks. Extensive simulations demonstrate that the accuracy of proposed KAQ framework outperforms traditional methods and successfully reduces the model's inference latency. 
\end{abstract}

\section{Introduction} \label{intro}

In the realm of wireless communications, the proliferation of edge computing presents new opportunities and challenges for efficient data processing~\cite{park2019edge,Elbamby2019edge,hu2020}. Edge computing allows for local data processing on devices such as user equipment or small base stations (BS), reducing latency and enhancing privacy, which is particularly crucial for real-time applications like mobile communications~\cite{yansha2025,bai2020,jeong2018}. A key task in these systems is multiple-input multiple-output (MIMO) detection~\cite{hu2023mimo,jin2025mimo,shao2021mimo}, which requires accurate and swift estimation of transmitted symbols from received signals. 

\begin{figure} [t]
	\centering 
		 \includegraphics[width=0.48\textwidth]{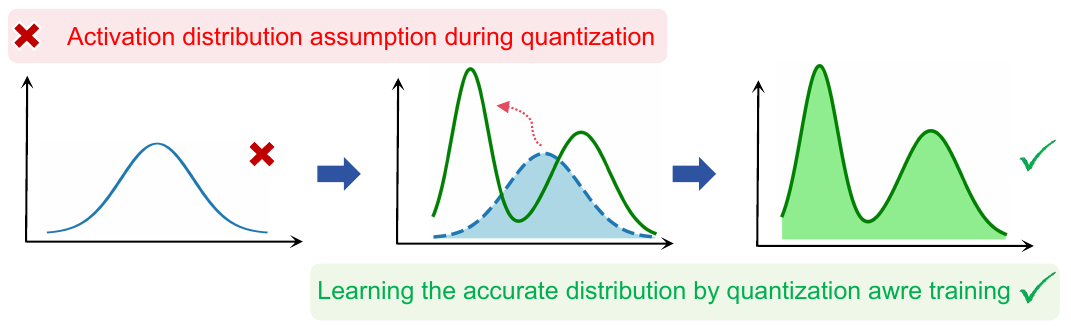}  
	
  	\caption{The process of increasing the awareness of accurate activation distribution for models}  \label{Fig1}
\end{figure}

Mathematically, two categories of optimization algorithms have been extensively applied to MIMO detection tasks and demonstrated to be highly effective: proximal gradient descent (PGD) algorithm~\cite{qingfeng2024,ren2025} and alternating direction method of multipliers (ADMM)~\cite{chang2016,ye2020} algorithm. PGD algorithm effectively handles non-smooth regularizers in MIMO detection by iteratively optimizing differentiable and proximal terms~\cite{ming2019}, while ADMM decouples complex constraints through variable splitting and dual updates, ensuring convergence under non-convex signal recovery settings, making them suitable for MIMO detection tasks. 

However, despite their theoretical merits, both PGD and ADMM algorithms face practical challenges in latency-sensitive BS deployments due to their iterative nature and computational overhead~\cite{fang2023,popo2018}. Proximal methods require multiple iterations to converge under non-smooth regularizers, while ADMM’s variable splitting and dual updates introduce coordination delays. These limitations challenge their direct applicability for practical adoption.

On the other hand, deep learning-based algorithms have become popular in communication research due to their low inference latency~\cite{barbarossa2023,niu2024info,xu2023,lei2024,Shang_Tang_Huang_Bi_He_Zhou_2019}. However, black box deep learning designs (e.g., multi-layer perceptrons (MLP) or convolutional neural networks~\cite{lei2022iccc}) do not incorporate the domain knowledge of communication systems. This usually results in unsatisfactory performance or the neural network not being able to converge during training.To address this, deep unfolding techniques have been employed to design efficient detection models~\cite{shuai2025,qftwc2024}, which transform iterative optimization algorithms into neural networks by regarding each iteration as one layer of the neural networks, offer a model-based deep learning paradigm for wireless communication tasks.

From the above arguments, applying deep unfolding models like PGD-Net~\cite{mou2022} and ADMM-Net~\cite{xiaodong2021} to the MIMO detection problem seems to be an obvious choice. However, these models typically employ full-precision floating-point operations, making it difficult to deploy on resource-constrained edge devices~\cite{anis2020jmlr,han2015deep}.

To address this, quantization is a promising way to reduce the complexity of model activations by converting high-bit floating-point data into lower-bit integer representations, preserving model functionality while minimizing computational overhead. 

Quantization methods can be broadly categorized into two types, post-training quantization (PTQ)~\cite{xiao2023,huang2025,huang2024} and quantization-aware training (QAT)~\cite{efficientqat}. PTQ quantizes a pre-trained model without further training, using a calibration dataset to determine quantization parameters. Although straightforward and computationally friendly for large models, PTQ often results in significant accuracy loss, particularly for small models like deep unfolding models. QAT, on the other hand, integrates quantization into the training process, using 'fake quantization' layers to simulate quantization effects~\cite{jacob2017,wuhai2023}. This allows the model to learn representations robust to quantization errors, generally leading to better accuracy than PTQ for deep unfolding~\cite{yonina2021}. However, traditional QAT methods rely on parametric distribution assumptions when designing the loss functions~\cite{gongiccv2019}, resulting in performance degradation, especially when the actual data distributions are nonparametric or vary across different parts of the network.

To address this limitation, this paper introduces a novel kernel-based adaptive quantization (KAQ) framework specifically designed for deep unfolding networks such as PGD-Net and ADMM-Net. KAQ is a quantization-aware training framework that leverages a joint kernel density estimation (KDE)~\cite{uncini2015,li2016nips} and maximum mean discrepancy (MMD)~\cite{nipsmmd2016} approach to align the activation distributions of the full-precision and quantized models during training. This approach maps the low-dimensional quantization discrepancies into a high-dimensional reproducing kernel Hilbert space (RKHS) to enable gradient-driven optimization by leveraging the kernel-induced feature representation~\cite{zhang2023nimt,zhang2024ntinr,hua2025}. By doing so, KAQ eliminates the need for prior distribution assumptions~\cite{zhang2023mint,zhang2025nonparametric}, leading to more accurate quantization.

Additionally, KAQ incorporates a dynamic step size updating that adjusts the quantization step size based on the signal-to-noise ratio (SNR). This adaptability ensures that the model is optimally quantized according to the channel condition of the communication system, minimizing information loss.

In the following Section~\ref{sys} and Section~\ref{method}, derivation and theoretical analysis focus on implementing the KAQ framework on the PGD-Net, while in Section~\ref{sim}. we also experimentally demonstrate the ADMM-Net quantization using KAQ framework.

Extensive simulations demonstrate that the proposed KAQ framework outperforms traditional quantization methods in terms of performance while successfully reducing the model's inference latency. This advancement paves the way for efficient and accurate deployment of deep unfolding networks on resource-constrained edge devices.

\section{System Model and Problem Formulation}   \label{sys}

\subsection{System Model}
We consider a multiple-input multiple-output (MIMO) communication system with \(N_t\) transmit antennas and \(N_r\) receive antennas. The received signal vector \(\mathbf{y} \in \mathbb{C}^{N_r}\) is modeled as
\begin{equation}
\mathbf{y} = \mathbf{H} \mathbf{x} + \mathbf{n},
\end{equation}
where \(\mathbf{x} \in \mathbb{C}^{N_t}\) denotes the transmitted signal vector, \(\mathbf{H} \in \mathbb{C}^{N_r \times N_t}\) represents the channel matrix, and \(\mathbf{n} \in \mathbb{C}^{N_r}\) is the additive white Gaussian noise (AWGN) vector with zero mean and covariance \(\sigma^2 \mathbf{I}\).

The entries of \(\mathbf{x}\) are assumed to be drawn from a finite modulation alphabet \(\mathcal{X}\), such as quadrature amplitude modulation (QAM). To enable the application of deep unfolding and the corresponding quantization techniques, we convert the complex-valued system into its real-valued equivalent. Specifically, we define
\begin{equation}
\tilde{\mathbf{y}} = \begin{bmatrix} \text{Re}(\mathbf{y}) \\ \text{Im}(\mathbf{y}) \end{bmatrix}, \quad
\tilde{\mathbf{x}} = \begin{bmatrix} \text{Re}(\mathbf{x}) \\ \text{Im}(\mathbf{x}) \end{bmatrix}, \nonumber
\end{equation}
\begin{equation}
\tilde{\mathbf{H}} = \begin{bmatrix} \text{Re}(\mathbf{H}) & -\text{Im}(\mathbf{H}) \\ \text{Im}(\mathbf{H}) & \text{Re}(\mathbf{H}) \end{bmatrix}, \quad
\tilde{\mathbf{n}} = \begin{bmatrix} \text{Re}(\mathbf{n}) \\ \text{Im}(\mathbf{n}) \end{bmatrix}.\nonumber
\end{equation}
This yields the real-valued system model
\begin{equation}
\tilde{\mathbf{y}} = \tilde{\mathbf{H}} \tilde{\mathbf{x}} + \tilde{\mathbf{n}},
\end{equation}
where \(\tilde{\mathbf{y}} \in \mathbb{R}^{2N_r}\), \(\tilde{\mathbf{x}} \in \mathbb{R}^{2N_t}\), \(\tilde{\mathbf{H}} \in \mathbb{R}^{2N_r \times 2N_t}\), and \(\tilde{\mathbf{n}} \in \mathbb{R}^{2N_r}\).

For notational simplicity, we omit the tildes in subsequent sections and refer to the real-valued system as
\begin{equation}
\mathbf{y} = \mathbf{H} \mathbf{x} + \mathbf{n},
\end{equation}
where \(\mathbf{y} \in \mathbb{R}^{M}\), \(\mathbf{x} \in \mathbb{R}^{N}\), \(\mathbf{H} \in \mathbb{R}^{M \times N}\), and \(\mathbf{n} \in \mathbb{R}^{M}\), with \(M = 2N_r\) and \(N = 2N_t\).

\begin{figure*} [t]
	\centering 
		 \includegraphics[width=0.68\textwidth]{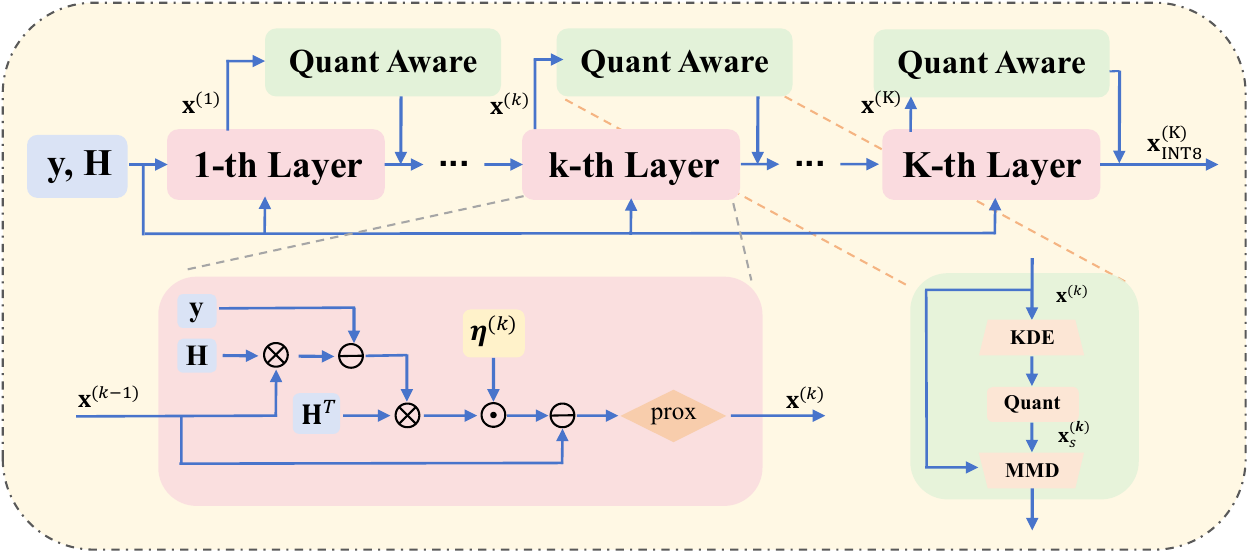}  
	
  	\caption{Architecture of the unfolded network, with details on the feed forward process of the k-th layer and the quantization aware module.}  \label{Fig2}
\end{figure*}

\subsection{Problem Formulation}
In MIMO detection, the objective is to recover the transmitted signal \(\mathbf{x}\) given the received signal \(\mathbf{y}\) and the known channel matrix \(\mathbf{H}\). The optimal maximum likelihood (ML) detector is formulated as
\begin{equation}
\hat{\mathbf{x}}_{\text{ML}} = \arg\min_{\mathbf{x} \in \mathcal{X}^{N}} \|\mathbf{y} - \mathbf{H} \mathbf{x}\|^2,
\end{equation}
where \(\mathcal{X}\) is the modulation alphabet. However, this combinatorial optimization problem is computationally prohibitive due to its exponential complexity with respect to \(N\).

To address this issue, we adopt a regularized least squares approach suitable for deep unfolding. We formulate the detection problem as
\begin{equation}
\min_{\mathbf{x} \in \mathbb{R}^{N}} \frac{1}{2} \|\mathbf{y} - \mathbf{H} \mathbf{x}\|^2 + \lambda \|\mathbf{x}\|_1,
\end{equation}
where \(\lambda > 0\) is a regularization parameter. 
This optimization problem can be efficiently solved using the PGD algorithm, which proceeds iteratively as follows:
\begin{align}
\mathbf{r}^{(k)} &= \mathbf{x}^{(k-1)} - \eta_k \mathbf{H}^T (\mathbf{H} \mathbf{x}^{(k-1)} - \mathbf{y}), \\
\mathbf{x}^{(k)} &= \text{prox}_{\lambda_k}(\mathbf{r}^{(k)}),
\end{align}
where $\mathbf{r}^{(k)}$ is the intermediate residual after the gradient update, \(\eta_k\) is the step size at the \(k\)-th iteration, $\mathbf{x}^{(k)}$ is the estimated transmitted signal at layer $k$, and \(\text{prox}_{\lambda_k}(\cdot)\) is the soft-thresholding operator, defined component-wise as
\begin{equation}
\left[ \text{prox}_{\lambda_k}(\mathbf{r}^{(k)}) \right]_n = \begin{cases}
r_n^{(k)} - \lambda_k & \text{if } r_n^{(k)} > \lambda_k, \\
0 & \text{if } |r_n^{(k)}| \leq \lambda_k, \\
r_n^{(k)} + \lambda_k & \text{if } r_n^{(k)} < -\lambda_k,
\end{cases}
\end{equation}
with \(\lambda_k\) serving as the threshold parameter and $r_n^{(k)}$ denotes the $n$-th element of $\mathbf{r}^{(k)}$ 


\begin{algorithm}[t]
\caption{KDE - MMD Quantization Aware Training}
\label{alg:training}
\begin{algorithmic}[1]
\STATE \textbf{Initialize} deep unfolding network with FP32 precision
\STATE \textbf{Collect} activation statistics $\{\mathbf{x}^{(k)}\}$ over training set
\FOR{each layer $k = 1$ to $K$}
    \STATE Kernel form calculation via KDE
    \STATE Initialize network parameters $\Theta^{(k)}$ and quantization steps $\Delta^{(k)}$
\ENDFOR
\WHILE{not converged}
    \STATE Forward pass: 
    \STATE \ \ \ Compute $\mathbf{x}_S^{(k)} = Q_b(\mathbf{x}^{(k)}; \Delta^{(k)})$
    \STATE \ \ \ Calculate $\mathcal{L}_{\text{total}}$ via~\eqref{eq:total_loss}
    \STATE Backward pass:
    \STATE \ \ \ Compute gradients $\nabla_{\Theta^{(k)}}\mathcal{L}_{\text{total}}$ via STE
    \STATE \ \ \ Update $\Theta^{(k)}$ and $\Delta^{(k)}$ using Adam optimizer
    \STATE Adjust $\Delta^{(k)}$ based on current SNR
\ENDWHILE
\end{algorithmic}
\end{algorithm}

\section{Proposed Method} \label{method}
The corresponding deep unfolding network of the PGD algorithm, termed PGD-Net, has demonstrated significant performance improvements in MIMO detection by incorporating learnable parameters to optimize iterative updates.

\subsection{PGD Deep Unfolding for MIMO Detection}
The transformation of the iterative PGD algorithm into a deep unfolded network (PGD-Net) involves systematically unrolling its update steps into a layered neural architecture, where each iteration corresponds to a dedicated network layer. Specifically, the gradient descent step 
\begin{align}
    \mathbf{r}^{(k)} &= \mathbf{x}^{(k-1)} - \eta_k \mathbf{H}^T (\mathbf{H} \mathbf{x}^{(k-1)} - \mathbf{y}) \nonumber
\end{align}
is implemented as a linear transformation layer with learnable step sizes $\{\eta_k\}$, while the proximal operator \(\text{prox}_{\lambda_k}(\cdot)\) interpreted as a sparsity-inducing activation function—is embedded as a nonlinear module with trainable thresholds $\{\lambda_k\}$. By parameterizing $\eta_k$ and $\lambda_k$ as layer-specific weights rather than fixed hyperparameters, PGD-Net enables end-to-end gradient-based optimization, allowing the network to adaptively refine both the gradient direction and sparsity constraints across iterations~\cite{mou2022}. Crucially, the intermediate residual $\mathbf{r}^{(k)}$
and signal estimate $\mathbf{x}^{(k)}$ are propagated sequentially through the layers, mimicking the iterative flow of PGD while integrating data-driven feature learning. 

However, deploying PGD-Net to resource-constrained environments, such as base stations and edge devices, faces computational bottlenecks due to its inherent complexity, necessitating efficient quantization strategies. While quantization-aware training (QAT) is conventionally tailored for lightweight models like deep unfolding networks~\cite{wuhai2023}, designing activation quantization loss functions remains challenging: conventional approaches relying on KL divergence or task-specific MSE often inadequately capture quantization discrepancies due to their reliance on restrictive distributional assumptions.

\subsection{KDE-MMD Loss}
To address this, we propose the KAQ framework. We first conducted a fit of the quantized activation values using a Gaussian kernel~\cite{tpami2015kernel,kim2023tpami}. By setting $\sigma^{(k)}$ within the Gaussian kernel as learnable parameters, we designed a loss function in conjunction with the MMD method. This enabled the distribution of the quantized activation values to progressively approach their true distribution during training, thereby substantially enhancing the accuracy of quantization - aware training.

The quantization is performed as:
\begin{equation}
\mathbf{x}_S^{(k)} = Q_b(\mathbf{x}^{(k)}), \quad Q_b(\mathbf{x}^{(k)}) = \Delta^{(k)} \cdot \text{round}\left(\frac{\mathbf{x}^{(k)}}{\Delta^{(k)}}\right), \label{eq:quantization}
\end{equation}
where \(b\) is the bit-width (8 in our case), and the initial step-size is:
\begin{equation}
\Delta^{(k)} = \frac{\max(|\mathbf{x}^{(k)}|)}{2^{b-1} - 1}. \label{eq:initial_step_size}
\end{equation}

To enable gradient-sensitive optimization that adaptively minimizes the divergence between full-precision and quantized model, we define the maximum mean discrepancy (MMD) loss at layer \(k\) as:
\begin{multline}
\mathcal{L}_{\text{MMD}}^{(k)} = \text{MMD}^2\left(\{\mathbf{x}_i^{(k)}\}_{i=1}^B, \{\mathbf{x}_{S,j}^{(k)}\}_{j=1}^B\right) \\
= \frac{1}{B^2} \sum_{i,j} \mathcal{K}(\mathbf{x}_i^{(k)}, \mathbf{x}_j^{(k)}) + \frac{1}{B^2} \sum_{i,j} \mathcal{K}(\mathbf{x}_{S,i}^{(k)}, \mathbf{x}_{S,j}^{(k)}) \\
- \frac{2}{B^2} \sum_{i,j} \mathcal{K}(\mathbf{x}_i^{(k)}, \mathbf{x}_{S,j}^{(k)}), \label{eq:mmd_loss}
\end{multline}
where \(B\) denotes the batch size. Taking the first term in the MMD loss as an example, 
\begin{equation}
\frac{1}{B^2} \sum_{i,j} \mathcal{K}(\mathbf{x}_i^{(k)}, \mathbf{x}_j^{(k)}) = \frac{1}{B^2} \sum_{i,j} \exp\left(-\frac{\|\mathbf{x}_i^{(k)} - \mathbf{x}_j^{(k)}\|^2}{2(\sigma_1^{(k)})^2}\right), \label{eq:kernel_term}
\end{equation}
where the Gaussian kernel $\mathcal{K}(\cdot,\cdot)$ is formulated by tow parts. From the distance metric perspective, $\|\mathbf{x}_i^{(k)} - \mathbf{x}_j^{(k)}\|^2$ computes the squared Euclidean distance between activation pairs from the full-precision model, measuring their pairwise dissimilarity. The adaptive bandwidth $\sigma_1^{(k)}$ serves as a data-driven scale parameter, for the second and third term of the MMD loss, the trainable parameters are demonstrated as $\sigma_2^{(k)}$ and $\sigma_3^{(k)}$. This method ensures that the model is able to learn the accurate activation distribution during training.

The gradients for quantized activations are computed as:
\begin{multline}
\frac{\partial \mathcal{L}_{\text{MMD}}}{\partial \mathbf{x}_{S,j}^{(k)}} = \frac{1}{B^2} \sum_{i=1}^B \frac{\partial \mathcal{K}(\mathbf{x}_{S,j}^{(k)}, \mathbf{x}_{S,i}^{(k)})}{\partial \mathbf{x}_{S,j}^{(k)}} \\- 
\frac{2}{B^2} \sum_{i=1}^B \frac{\partial \mathcal{K}(\mathbf{x}_{S,j}^{(k)}, \mathbf{x}_i^{(k)})}{\partial \mathbf{x}_{S,j}^{(k)}}, \label{eq:mmd_gradient}
\end{multline}
taking the first term as an example, with the kernel gradient derivative:
\begin{equation}
\frac{\partial \mathcal{K}(\mathbf{x}_{S,j}^{(k)}, \mathbf{x}_{S,i}^{(k)})}{\partial \mathbf{x}_{S,j}^{(k)}} = -\frac{\mathbf{x}_{S,j}^{(k)} - \mathbf{x}_{S,i}^{(k)}}{(\sigma_{2}^{(k)})^2} \mathcal{K}(\mathbf{x}_{S,j}^{(k)}, \mathbf{x}_{S,i}^{(k)}). \label{eq:kernel_derivative}
\end{equation}
This approach mitigates the accuracy loss using traditional QAT. Comparing to simply using MSE as the loss function during training, this kernel-based MMD approach bridges the gaps of activation distribution between the quantized model and full precisions. The specific structure of the quantization aware module and its integration with the unfolded network are illustrated in Fig.~\ref{Fig2}.

\begin{figure} [t]
	\centering 
		 \includegraphics[width=0.4\textwidth]{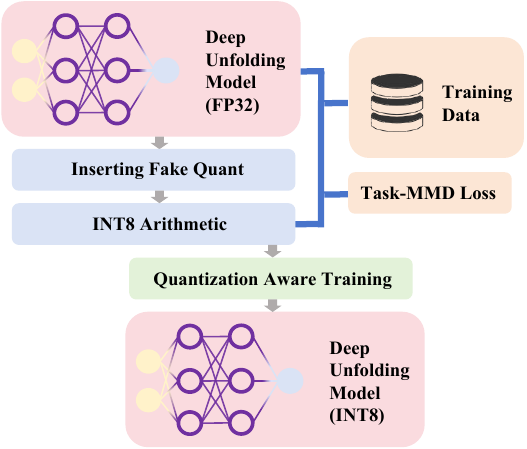}  
	
  	\caption{The KAQ training scheme from full precision model to quantized version}  \label{Fig3}
\end{figure}

\subsection{Joint Optimization of Quantization Step-Size}
The time-varying nature of MIMO channels also requires adaptive quantization strategies. Static quantization steps fail to adapt different SNR conditions. Given the dynamic channel conditions in MIMO systems, we jointly optimize \(\Delta^{(k)}\) based on SNR, formulated as:
\begin{equation}
\Delta^{(k)} = f_\theta(\text{SNR}) = \alpha \cdot \text{SNR}^{-1/2} + \gamma, \label{eq:step_size_optimization}
\end{equation}
where the coefficients $\alpha$, and $\gamma$ are learnable parameters. This design captures that the $\text{SNR}^{-1/2}$ term reduces quantization step-sizes under high-noise conditions to preserve signal details. And the bias term $\gamma$ maintains minimum quantization resolution even under ideal channel conditions.

\subsection{Training with Quantization-aware Loss}
The total loss combines the mean squared error (MSE) and the MMD loss, optimizing the detection accuracy and capturing the matching distribution of the activations at the same time,
\begin{equation}
\mathcal{L}_{\text{total}} = \underbrace{\frac{1}{B} \sum_{i=1}^B \| \mathbf{x}_{\text{true}} - \mathbf{x}_S^{(K)} \|_2^2}_{\text{Detection accuracy}} + \underbrace{\epsilon \sum_{k=1}^K \mathcal{L}_{\text{MMD}}^{(k)}}_{\text{Distribution matching}}, 
\label{eq:total_loss}
\end{equation}
where \(B\) is the batch size, and \(\epsilon\) balances the terms. Network parameters $\Theta^{(k)}$ and $\Delta^{(k)}$ are updated via gradient descent:
\begin{align}
\Theta^{(k)} &\leftarrow \Theta^{(k)} - \mu \nabla_{\Theta_S} \mathcal{L}_{\text{total}}, \label{eq:parameter_update} \\
\Delta^{(k)} &\leftarrow \Delta^{(k)} - \mu \nabla_{\Delta^{(k)}} \mathcal{L}_{\text{total}}, \label{eq:step_size_update}
\end{align}
where \(\mu\) is the learning rate, enabling end-to-end quantization-aware training.
For gradient propagation during training, we use the straight-through estimator (STE):
\begin{equation}
\frac{\partial Q_b(x)}{\partial x} = 
\begin{cases} 
1 & \text{if } x \in [-\Delta^{(k)}(2^{b-1}-1), \Delta^{(k)}(2^{b-1}-1)], \\
0 & \text{otherwise},
\end{cases} \label{eq:ste_gradient}
\end{equation}
which preserves gradient magnitude within the active quantization range while enabling quantization-aware training. Algorithm~\ref{alg:training} summarizes the KAQ training approach while Fig.~\ref{Fig3} illustrates the whole training scheme.

\begin{figure} [t]
	\centering
 	{\label{fig_ber} 
		\includegraphics[width=0.4\textwidth]{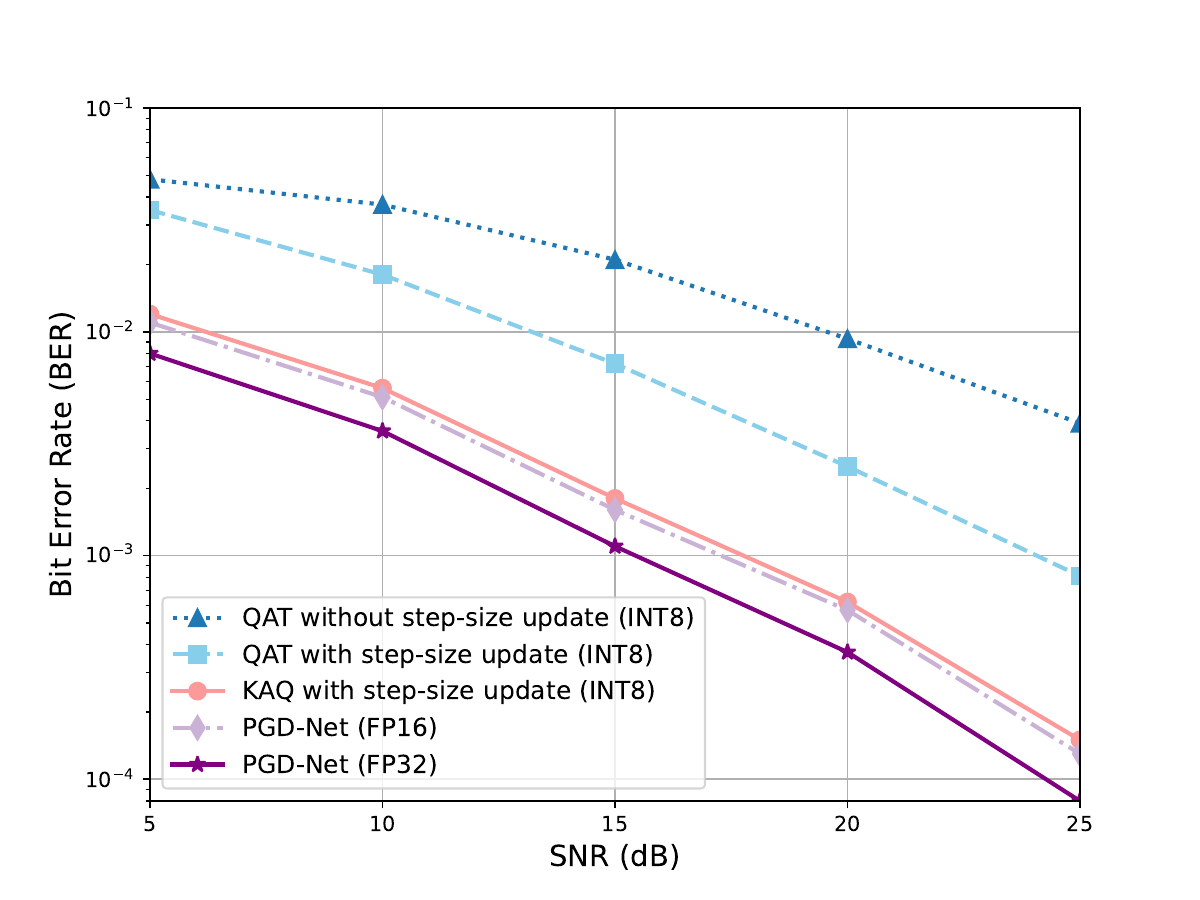}}
	  
  	\caption{BER versus SNR.}  \label{fig_ber}
\end{figure}

\section{Simulation Results} \label{sim}

To assess the efficacy of the proposed KAQ framework for MIMO detection, numerical experiments are conducted at diverse angles. We compare the performance of quantized PGD-Net and ADMM-Net with and without the proposed KAQ framework. Furthermore, we compare the inference latency and memory storage of the quantized deep unfolding models with the full precision models.

The simulation setup employs a 16$\times$16 MIMO system ($N_t = 16$ transmit and $N_r = 16$ receive antennas), corresponding to a real-valued system of dimensions $N = 32$ and $M = 32$. Transmitted symbols are generated from a 16-QAM constellation with independent real and imaginary components drawn from $\mathcal{X} = \{\pm1, \pm3\}$. The Rayleigh fading channel matrix $\mathbf{H}$ is configured with entries independently sampled from $\mathcal{N}(0, 1/N_r)$, while additive noise $\mathbf{n}$ is introduced at SNR levels ranging from 0 dB to 25 dB in 5-dB increments. This setup enables comprehensive evaluation under varying channel realizations and noise conditions.

The training protocol utilizes a dataset of $5 \times10^4$ samples containing channel matrices $\mathbf{H}$, received signals $\mathbf{y}$, and transmitted symbols $\mathbf{x}$. Both PGD-Net and ADMM-Net architectures are unfolded into $K = 5$ trainable layers, initialized with uniform parameters $\eta_k = 0.1$ and $\lambda_k = 0.05$ for all layers $k$. Training is performed using the Adam optimizer with an initial learning rate of $10^{-3}$, dynamically reduced by half every 10 epochs over a total of 50 training epochs. Mini-batches of size 128 are employed to balance computational efficiency and gradient estimation accuracy. 

\subsection{Model Accuracy}
\label{experiment_accuracy}
From the perspective of model accuracy, we first compare the bit error rate (BER) versus SNR of PGD-Net under different precisions. As shown in Fig.~\ref{fig_ber}, the experimental results demonstrate that dynamically adjusting the quantization step size based on instantaneous SNR enables the quantized PGD-Net to achieve lower BER. When implementing the MMD loss, the incorporation of Gaussian kernels and learnable parameters $\sigma^{(k)}$ provides distinct advantages over conventional QAT approaches that solely employ MSE loss. Specifically, the MMD loss effectively captures distributional discrepancies induced by model quantization through its kernel-based metric. During the parameter update process of $\sigma^{(k)}$, the activation distributions progressively approximate the true distribution starting from an initial Gaussian assumption. This adaptive mechanism allows the KAQ dynamic quantization method to perform nearly as well as the FP16-precision PGD-Net baseline. At the same time, it significantly surpasses the traditional QAT method that depend on MSE loss optimization. In Fig.~\ref{fig_accuracy}, it also can be observed that implementing the KAQ training scheme to quantize the ADMM-Net, can converge to a higher accuracy level compared to the traditional QAT method.

\begin{figure} [t]
	\centering
 	{\label{fig_accuracy} 
		\includegraphics[width=0.4\textwidth]{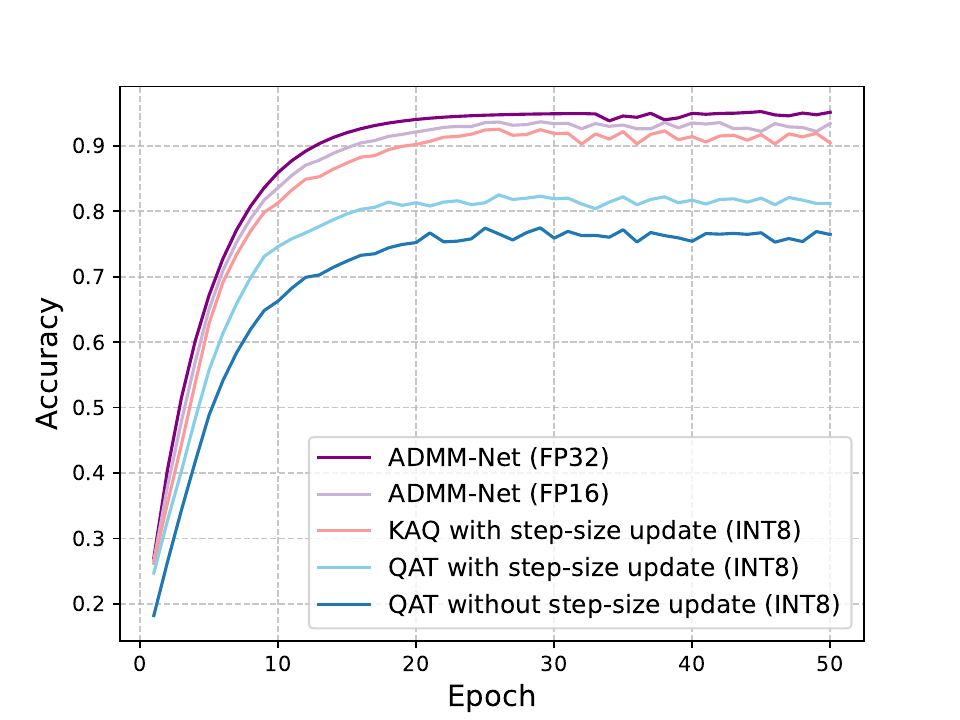}}
	  
  	\caption{Model accuracy versus training epoch}  \label{fig_accuracy}
\end{figure}

\subsection{Inference Latency}
\begin{figure} [t]
	\centering
 	{\label{fig6} 
		\includegraphics[width=0.40\textwidth]{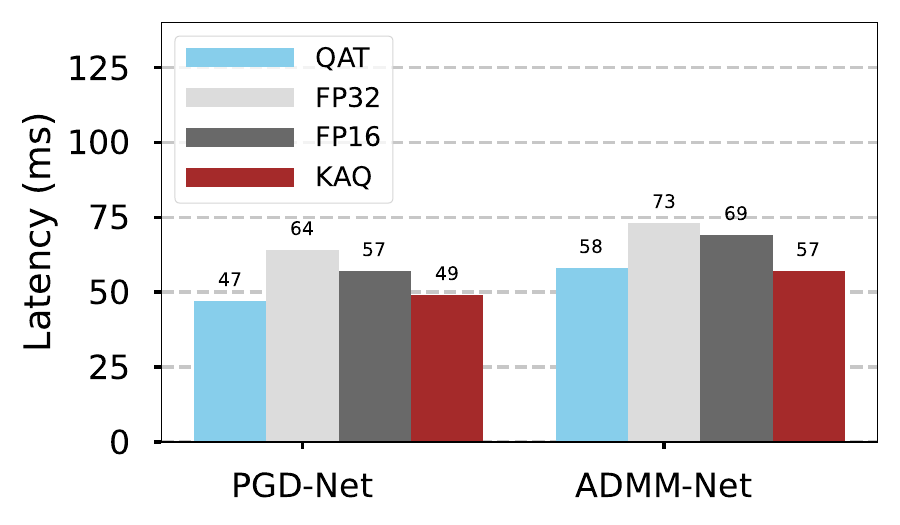}}
	  
  	\caption{Comapision of inference latency}  \label{fig_latency}
\end{figure}

Next, we illustrate the inference latency of both PGD-Net and ADMM-Net under different precisions. As shown in Fig.~\ref{fig_latency}, the inference latency of the quantized models using KAQ is reduced in 20\% compared to the full precision version. While keeping at the same level of inference latency with traditional QAT method, KAQ significantly outperforms on the model accuracy according to Section~\ref{experiment_accuracy}.

\section{Conclusions}

In this paper, we proposed a kernel-based adaptive quantization approach for model-driven deep unfolding networks. Specifically, we used a learnable Gaussian kernel that cooperated with the MMD approach to align the quantization-aware training loss. By further intrduced the dynamic quantization step-size, the proposed KAQ framework can be adaptive to wireless communication tasks based on the SNR conditions. Simulation results showed that the kernel-based adaptive quantization method achieves better accuracy performance than traditional QAT methods and efficiently reduces the inference time compares to the full precision models.

\section*{Impact Statement} 

Recent interest in efficient model deployment and edge computing has grown significantly, especially for deep unfolding models, driven by the ability of mathematical guidance. This work focuses on improving the accuracy and reducing the inference latency while implementing quantization through a novel kernel-based perspective, which has the potential to positively impact communication society. Furthermore, it connects model-driven deep learning with kernel-based learning, also has the potential on expanding its application in various types of  deep neural networks. 

\bibliography{ref}

\begin{thebibliography}{46}
\providecommand{\natexlab}[1]{#1}
\providecommand{\url}[1]{\texttt{#1}}
\expandafter\ifx\csname urlstyle\endcsname\relax
  \providecommand{\doi}[1]{doi: #1}\else
  \providecommand{\doi}{doi: \begingroup \urlstyle{rm}\Url}\fi

\bibitem[Bai et~al.(2020)Bai, Pan, Deng, Elkashlan, Nallanathan, and
  Hanzo]{bai2020}
Bai, T., Pan, C., Deng, Y., Elkashlan, M., Nallanathan, A., and Hanzo, L.
\newblock Latency minimization for intelligent reflecting surface aided mobile
  edge computing.
\newblock \emph{IEEE Journal on Selected Areas in Communications}, 2020.

\bibitem[Barbarossa et~al.(2023)Barbarossa, Comminiello, Grassucci, Pezone,
  Sardellitti, and Di~Lorenzo]{barbarossa2023}
Barbarossa, S., Comminiello, D., Grassucci, E., Pezone, F., Sardellitti, S.,
  and Di~Lorenzo, P.
\newblock Semantic communications based on adaptive generative models and
  information bottleneck.
\newblock \emph{IEEE Communications Magazine}, 2023.

\bibitem[Chang et~al.(2016)Chang, Hong, Liao, and Wang]{chang2016}
Chang, T.-H., Hong, M., Liao, W.-C., and Wang, X.
\newblock Asynchronous distributed admm for large-scale optimization—part i:
  Algorithm and convergence analysis.
\newblock \emph{IEEE Transactions on Signal Processing}, 2016.

\bibitem[Chen et~al.(2024)Chen, Shao, Xu, Wang, Gao, Zhang, Qiao, and
  Luo]{efficientqat}
Chen, M., Shao, W., Xu, P., Wang, J., Gao, P., Zhang, K., Qiao, Y., and Luo, P.
\newblock Efficientqat: Efficient quantization-aware training for large
  language models.
\newblock \emph{arXiv preprint arXiv:2407.11062}, 2024.

\bibitem[Elbamby et~al.(2019)Elbamby, Perfecto, Liu, Park, Samarakoon, Chen,
  and Bennis]{Elbamby2019edge}
Elbamby, M.~S., Perfecto, C., Liu, C.-F., Park, J., Samarakoon, S., Chen, X.,
  and Bennis, M.
\newblock Wireless edge computing with latency and reliability guarantees.
\newblock \emph{Proceedings of the IEEE}, 2019.

\bibitem[Elgabli et~al.(2020)Elgabli, Park, Bedi, Bennis, and
  Aggarwal]{anis2020jmlr}
Elgabli, A., Park, J., Bedi, A.~S., Bennis, M., and Aggarwal, V.
\newblock Gadmm: Fast and communication efficient framework for distributed
  machine learning.
\newblock \emph{Journal of Machine Learning Research}, 2020.

\bibitem[Gong et~al.(2019)Gong, Liu, Jiang, Li, Hu, Lin, Yu, and
  Yan]{gongiccv2019}
Gong, R., Liu, X., Jiang, S., Li, T., Hu, P., Lin, J., Yu, F., and Yan, J.
\newblock Differentiable soft quantization: Bridging full-precision and low-bit
  neural networks.
\newblock In \emph{ICCV}, 2019.

\bibitem[Han et~al.(2025)Han, Wang, Wang, Zhang, Chen, Lin, Li, Xu, Eldar, Hao,
  and Pan]{shuai2025}
Han, R., Wang, S., Wang, S., Zhang, Z., Chen, J., Lin, S., Li, C., Xu, C.,
  Eldar, Y.~C., Hao, Q., and Pan, J.
\newblock Neupan: Direct point robot navigation with end-to-end model-based
  learning.
\newblock \emph{IEEE Transactions on Robotics}, 2025.

\bibitem[Han et~al.(2016)Han, Mao, and Dally]{han2015deep}
Han, S., Mao, H., and Dally, W.~J.
\newblock Deep compression: Compressing deep neural networks with pruning,
  trained quantization and huffman coding.
\newblock In \emph{ICLR}, 2016.

\bibitem[Hu et~al.(2020)Hu, Deng, Saad, Bennis, and Aghvami]{hu2020}
Hu, F., Deng, Y., Saad, W., Bennis, M., and Aghvami, A.~H.
\newblock Cellular-connected wireless virtual reality: Requirements,
  challenges, and solutions.
\newblock \emph{IEEE Communications Magazine}, 2020.

\bibitem[Hu et~al.(2023)Hu, Gao, Zhang, Li, and Xu]{hu2023mimo}
Hu, Q., Gao, F., Zhang, H., Li, G.~Y., and Xu, Z.
\newblock Understanding deep mimo detection.
\newblock \emph{IEEE Transactions on Wireless Communications}, 2023.

\bibitem[Hua et~al.(2025)Hua, Jiang, Lv, Zhang, Ding, Sun, Qi, Fan, Zhu, and
  Zhou]{hua2025}
Hua, E., Jiang, C., Lv, X., Zhang, K., Ding, N., Sun, Y., Qi, B., Fan, Y., Zhu,
  X., and Zhou, B.
\newblock Fourier position embedding: Enhancing attention's periodic extension
  for length generalization.
\newblock In \emph{ICML}, 2025.

\bibitem[Huang et~al.(2024)Huang, Liu, Qin, Li, Zhang, Liu, Magno, and
  Qi]{huang2024}
Huang, W., Liu, Y., Qin, H., Li, Y., Zhang, S., Liu, X., Magno, M., and Qi, X.
\newblock Billm: Pushing the limit of post-training quantization for llms.
\newblock In \emph{ICML}, 2024.

\bibitem[Huang et~al.(2025)Huang, Qin, Liu, Li, Liu, Benini, Magno, and
  Qi]{huang2025}
Huang, W., Qin, H., Liu, Y., Li, Y., Liu, X., Benini, L., Magno, M., and Qi, X.
\newblock Slim-llm: Salience-driven mixed-precision quantization for large
  language models.
\newblock In \emph{ICML}, 2025.

\bibitem[Jacob et~al.(2018)Jacob, Kligys, Chen, Zhu, Tang, Howard, Adam, and
  Kalenichenko]{jacob2017}
Jacob, B., Kligys, S., Chen, B., Zhu, M., Tang, M., Howard, A., Adam, H., and
  Kalenichenko, D.
\newblock Quantization and training of neural networks for efficient
  integer-arithmetic-only inference.
\newblock In \emph{CVPR}, 2018.

\bibitem[Jayasumana et~al.(2015)Jayasumana, Hartley, Salzmann, Li, and
  Harandi]{tpami2015kernel}
Jayasumana, S., Hartley, R., Salzmann, M., Li, H., and Harandi, M.
\newblock Kernel methods on riemannian manifolds with gaussian rbf kernels.
\newblock \emph{IEEE Transactions on Pattern Analysis and Machine
  Intelligence}, 2015.

\bibitem[Jeong et~al.(2018)Jeong, Oh, Kim, Park, Bennis, and Kim]{jeong2018}
Jeong, E., Oh, S., Kim, H., Park, J., Bennis, M., and Kim, S.-L.
\newblock Communication-efficient on-device machine learning: Federated
  distillation and augmentation under non-iid private data.
\newblock In \emph{NeurIPS Workshop on Machine Learning on the Phone and other
  Consumer Devices (MLPCD)}, 2018.

\bibitem[Johnston et~al.(2021)Johnston, Li, Lops, and Wang]{xiaodong2021}
Johnston, J., Li, Y., Lops, M., and Wang, X.
\newblock Admm-net for communication interference removal in stepped-frequency
  radar.
\newblock \emph{IEEE Transactions on Signal Processing}, 2021.

\bibitem[Khobahi et~al.(2021)Khobahi, Shlezinger, Soltanalian, and
  Eldar]{yonina2021}
Khobahi, S., Shlezinger, N., Soltanalian, M., and Eldar, Y.~C.
\newblock Lord-net: Unfolded deep detection network with low-resolution
  receivers.
\newblock \emph{IEEE Transactions on Signal Processing}, 2021.

\bibitem[Kim et~al.(2023)Kim, Lee, and Liang]{kim2023tpami}
Kim, J., Lee, Y., and Liang, Z.
\newblock The geometry of nonlinear embeddings in kernel discriminant analysis.
\newblock \emph{IEEE Transactions on Pattern Analysis and Machine
  Intelligence}, 2023.

\bibitem[Lei et~al.(2022)Lei, Zhou, Jian, Liu, Zhang, Feng, and
  Wu]{lei2022iccc}
Lei, J., Zhou, J., Jian, Z., Liu, H., Zhang, L., Feng, Y., and Wu, Z.
\newblock A wiener filter denoising based intelligent modulation recognition
  system.
\newblock In \emph{ICCC}, 2022.

\bibitem[Lei et~al.(2024)Lei, Li, Yung, Leng, Lin, and Wu]{lei2024}
Lei, J., Li, Y., Yung, L.-Y., Leng, Y., Lin, Q., and Wu, Y.-C.
\newblock Understanding complex-valued transformer for modulation recognition.
\newblock \emph{IEEE Wireless Communications Letters}, 2024.

\bibitem[Li et~al.(2016)Li, Yang, and Wong]{li2016nips}
Li, D., Yang, K., and Wong, W.~H.
\newblock Density estimation via discrepancy based adaptive sequential
  partition.
\newblock In \emph{NeurIPS}, 2016.

\bibitem[Li et~al.(2019)Li, Shi, and Yan]{ming2019}
Li, Z., Shi, W., and Yan, M.
\newblock A decentralized proximal-gradient method with network independent
  step-sizes and separated convergence rates.
\newblock \emph{IEEE Transactions on Signal Processing}, 2019.

\bibitem[Lin et~al.(2024{\natexlab{a}})Lin, Li, Kou, Chang, and Wu]{qftwc2024}
Lin, Q., Li, Y., Kou, W.-B., Chang, T.-H., and Wu, Y.-C.
\newblock Communication-efficient activity detection for cell-free massive
  mimo: An augmented model-driven end-to-end learning framework.
\newblock \emph{IEEE Transactions on Wireless Communications},
  2024{\natexlab{a}}.

\bibitem[Lin et~al.(2024{\natexlab{b}})Lin, Li, Wu, and Zhang]{qingfeng2024}
Lin, Q., Li, Y., Wu, Y.-C., and Zhang, R.
\newblock Intelligent reflecting surface aided activity detection for massive
  access: Performance analysis and learning approach.
\newblock \emph{IEEE Transactions on Wireless Communications},
  2024{\natexlab{b}}.

\bibitem[Mou et~al.(2022)Mou, Wang, and Zhang]{mou2022}
Mou, C., Wang, Q., and Zhang, J.
\newblock Deep generalized unfolding networks for image restoration.
\newblock In \emph{CVPR}, 2022.

\bibitem[Park et~al.(2019)Park, Samarakoon, Bennis, and Debbah]{park2019edge}
Park, J., Samarakoon, S., Bennis, M., and Debbah, M.
\newblock Wireless network intelligence at the edge.
\newblock \emph{Proceedings of the IEEE}, 2019.

\bibitem[Popovski et~al.(2018)Popovski, Nielsen, Stefanovic, Carvalho, Strom,
  Trillingsgaard, Bana, Kim, Kotaba, Park, and Sorensen]{popo2018}
Popovski, P., Nielsen, J.~J., Stefanovic, C., Carvalho, E.~d., Strom, E.,
  Trillingsgaard, K.~F., Bana, A.-S., Kim, D.~M., Kotaba, R., Park, J., and
  Sorensen, R.~B.
\newblock Wireless access for ultra-reliable low-latency communication:
  Principles and building blocks.
\newblock \emph{IEEE Network}, 2018.

\bibitem[Ren et~al.(2025)Ren, Lin, Lei, Li, and Wu]{ren2025}
Ren, Z., Lin, Q., Lei, J., Li, Y., and Wu, Y.-C.
\newblock Mixture of experts-augmented deep unfolding for activity detection in
  irs-aided systems.
\newblock \emph{arXiv}, 2025.

\bibitem[Scardapane et~al.(2015)Scardapane, Comminiello, Scarpiniti, and
  Uncini]{uncini2015}
Scardapane, S., Comminiello, D., Scarpiniti, M., and Uncini, A.
\newblock Online sequential extreme learning machine with kernels.
\newblock \emph{IEEE Transactions on Neural Networks and Learning Systems},
  2015.

\bibitem[Shang et~al.(2019)Shang, Tang, Huang, Bi, He, and
  Zhou]{Shang_Tang_Huang_Bi_He_Zhou_2019}
Shang, C., Tang, Y., Huang, J., Bi, J., He, X., and Zhou, B.
\newblock End-to-end structure-aware convolutional networks for knowledge base
  completion.
\newblock In \emph{Proceedings of the AAAI Conference on Artificial
  Intelligence}, 2019.

\bibitem[Shao \& Ma(2021)Shao and Ma]{shao2021mimo}
Shao, M. and Ma, W.-K.
\newblock Binary mimo detection via homotopy optimization and its deep
  adaptation.
\newblock \emph{IEEE Transactions on Signal Processing}, 2021.

\bibitem[Tolstikhin et~al.(2016)Tolstikhin, Sriperumbudur, and
  Sch\"{o}lkopf]{nipsmmd2016}
Tolstikhin, I., Sriperumbudur, B.~K., and Sch\"{o}lkopf, B.
\newblock Minimax estimation of maximum mean discrepancy with radial kernels.
\newblock In \emph{NeurIPS}, 2016.

\bibitem[Wang et~al.(2025)Wang, Pan, Huang, Jin, and Caire]{jin2025mimo}
Wang, Z., Pan, C., Huang, Y., Jin, S., and Caire, G.
\newblock Randomized iterative algorithms for distributed massive mimo
  detection.
\newblock \emph{IEEE Transactions on Signal Processing}, 2025.

\bibitem[Wu et~al.(2023)Wu, He, Tan, Qi, and Huang]{wuhai2023}
Wu, H., He, R., Tan, H., Qi, X., and Huang, K.
\newblock Vertical layering of quantized neural networks for heterogeneous
  inference.
\newblock \emph{IEEE Transactions on Pattern Analysis and Machine
  Intelligence}, 45, 2023.

\bibitem[Xiao et~al.(2023)Xiao, Lin, Seznec, Wu, Demouth, and Han]{xiao2023}
Xiao, G., Lin, J., Seznec, M., Wu, H., Demouth, J., and Han, S.
\newblock Smoothquant: Accurate and efficient post-training quantization for
  large language models.
\newblock In \emph{ICML}, 2023.

\bibitem[Xu et~al.(2023)Xu, Zhou, and Deng]{xu2023}
Xu, Y., Zhou, H., and Deng, Y.
\newblock Task-oriented semantics-aware communication for wireless uav control
  and command transmission.
\newblock \emph{IEEE Communications Letters}, 2023.

\bibitem[Ye et~al.(2020)Ye, Chen, Xiao, Skoglund, and Vincent~Poor]{ye2020}
Ye, Y., Chen, H., Xiao, M., Skoglund, M., and Vincent~Poor, H.
\newblock Privacy-preserving incremental admm for decentralized consensus
  optimization.
\newblock \emph{IEEE Transactions on Signal Processing}, 2020.

\bibitem[Yilmaz et~al.(2024)Yilmaz, Niu, Bai, Han, Deng, and
  Gündüz]{niu2024info}
Yilmaz, S.~F., Niu, X., Bai, B., Han, W., Deng, L., and Gündüz, D.
\newblock High perceptual quality wireless image delivery with denoising
  diffusion models.
\newblock In \emph{INFOCOM WKSHPS}, 2024.

\bibitem[Yu et~al.(2023)Yu, Zhou, Chen, Li, and Dong]{fang2023}
Yu, J., Zhou, R., Chen, C., Li, B., and Dong, F.
\newblock Asfl: Adaptive semi-asynchronous federated learning for balancing
  model accuracy and total latency in mobile edge networks.
\newblock In \emph{ICPP}, 2023.

\bibitem[Zhang et~al.(2023{\natexlab{a}})Zhang, Cao, Liu, Tsang, and
  Kwok]{zhang2023mint}
Zhang, C., Cao, X., Liu, W., Tsang, I., and Kwok, J.
\newblock Nonparametric teaching for multiple learners.
\newblock In \emph{NeurIPS}, 2023{\natexlab{a}}.

\bibitem[Zhang et~al.(2023{\natexlab{b}})Zhang, Cao, Liu, Tsang, and
  Kwok]{zhang2023nimt}
Zhang, C., Cao, X., Liu, W., Tsang, I., and Kwok, J.
\newblock Nonparametric iterative machine teaching.
\newblock In \emph{ICML}, 2023{\natexlab{b}}.

\bibitem[Zhang et~al.(2024)Zhang, Luo, Li, Wu, and Wong]{zhang2024ntinr}
Zhang, C., Luo, S., Li, J., Wu, Y.-C., and Wong, N.
\newblock Nonparametric teaching of implicit neural representations.
\newblock In \emph{ICML}, 2024.

\bibitem[Zhang et~al.(2025)Zhang, Bu, Ren, Liu, Wu, and
  Wong]{zhang2025nonparametric}
Zhang, C., Bu, W., Ren, Z., Liu, Z., Wu, Y.-C., and Wong, N.
\newblock Nonparametric teaching for graph property learners.
\newblock In \emph{ICML}, 2025.

\bibitem[Zhao et~al.(2025)Zhao, Wang, Chu, Song, Deng, Nallanathan, and
  Karagiannidis]{yansha2025}
Zhao, L., Wang, Y., Chu, X., Song, S., Deng, Y., Nallanathan, A., and
  Karagiannidis, G.~K.
\newblock Open-source edge ai for 6g wireless networks.
\newblock \emph{IEEE Network}, 2025.

\end{thebibliography}
\bibliographystyle{icml2025}



\end{document}